\pdfoutput=1

\documentclass[11pt]{article}

\usepackage{acl}

\usepackage{times}
\usepackage{latexsym}
\usepackage{bbding}

\usepackage[T1]{fontenc}

\usepackage[utf8]{inputenc}

\usepackage{microtype}
\usepackage{paralist}

\usepackage{graphicx}
\usepackage{multirow}
\usepackage{color}

\title{EasyECR: A Library for Easy Implementation and Evaluation of Event Coreference Resolution Models}

\author{Yuncong Li\textsuperscript{\rm $\spadesuit$},~
  Tianhua Xu\textsuperscript{\rm $\diamondsuit$}\footnotemark[1]\;,~Sheng-hua Zhong\textsuperscript{\rm $\diamondsuit$},~Haiqin Yang\textsuperscript{\rm $\spadesuit$}\footnotemark[2]\\
  \textsuperscript{\rm $\spadesuit$}International Digital Economy Academy (IDEA), China\\
\textsuperscript{\rm $\diamondsuit$}Shenzhen University\\
\texttt{hqyang@ieee.org}  
}

\begin{document}
\maketitle
\renewcommand{\thefootnote}{\fnsymbol{footnote}} 
\footnotetext[1]{Work done when Tianhua was interned at IDEA.  } 
\footnotetext[2]{The corresponding author.} 

\begin{abstract}
Event Coreference Resolution (ECR) is the task of clustering event mentions that refer to the same real-world event. Despite significant advancements, ECR research faces two main challenges: limited generalizability across domains due to narrow dataset evaluations, and difficulties in comparing models within diverse ECR pipelines. To address these issues, we develop EasyECR, the first open-source library designed to standardize data structures and abstract ECR pipelines for easy implementation and fair evaluation. More specifically, EasyECR integrates seven representative pipelines and ten popular benchmark datasets, enabling model evaluations on various datasets and promoting the development of robust ECR pipelines. By conducting extensive evaluation via our EasyECR, we find that, \lowercase\expandafter{\romannumeral1}) the representative ECR pipelines cannot generalize across multiple datasets, hence evaluating ECR pipelines on multiple datasets is necessary, \lowercase\expandafter{\romannumeral2}) all models in ECR pipelines have a great effect on pipeline performance, therefore, when one model in ECR pipelines are compared, it is essential to ensure that the other models remain consistent. Additionally, reproducing ECR results is not trivial, and the developed library can help reduce this discrepancy. The experimental results provide valuable baselines for future research. 

\end{abstract}

\section{Introduction}

\begin{figure}
	\centering
	\includegraphics[scale=0.46]{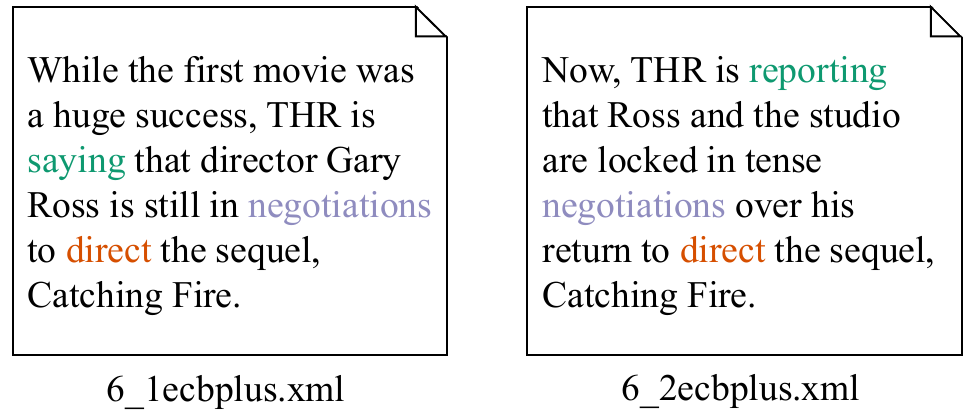}
	\caption{An example of event coreference resolution, which contains three coreferential chains from two documents: \{\textit{saying}, \textit{reporting}\}, \{\textit{negotiations}, \textit{negotiations}\} and \{\textit{direct}, \textit{direct}\}.}
	\label{fig:example}
\end{figure}

Event coreference resolution (ECR) plays a pivotal role in event analysis by identifying mentions of the same event~\citep{lu2018event,liu2023brief}.  It encompasses two types based on document analysis: within a single document, referred to as ``within-document coreference resolution'' (WDCR), and across different documents, known as ``cross-document coreference resolution'' (CDCR).  Figure~\ref{fig:example} illustrates an example of ECR, featuring three coreferential chains comprising six event mentions extracted from two documents.

Recently, ECR has witnessed substantial progress, with numerous proposed models shaping its development. Research paradigms for the ECR task can be categorized into four main groups according to technology iterations: rule-based, feature-based, transformer-based, and large language model (LLM)-based approaches. Rule-based approaches rely on a set of predefined rules for prediction. These rules are hand-designed by human experts based on domain-specific knowledge and may include syntactic rules or combinations of specific conditions. Notable contributions in this field include~\citep{araki-mitamura-2015-joint, peng-etal-2016-event}. Feature-based methods focus on utilizing lexical, syntactic, and semantic information in text. Employing established machine learning frameworks such as decision trees and conditional random fields, these methods have been refined by a host of researchers, including~\citep{yang-etal-2015-hierarchical, lu-etal-2016-joint, liu-etal-2018-graph, choubey-huang-2018-improving, Tran2021exploiting, choubey-huang-2017-event, lee-etal-2017-end, hsu-horwood-2022-contrastive}. Transformers-based methods provide powerful representations of text that capture both local and long-range dependencies, making them widely adopted in ECR tasks. Representative works in this category include~\citep{held-etal-2021-focus, kriman-ji-2021-joint, lu-ng-2021-conundrums, lu-ng-2021-constrained, zeng-etal-2020-event, allaway-etal-2021-sequential, caciularu-etal-2021-cdlm-cross, barhom-etal-2019-revisiting, ahmed-etal-2023-2}. In the most recent wave of innovation, LLMs have emerged as a formidable force in ECR research. Pioneering studies by~\citep{xu-etal-2023-corefprompt, ravi-etal-2023-happens} have set the stage for a new era of exploration and discovery.


\begin{table}
\centering
\begin{tabular}{l|l}
\hline
Paper                                             & Datasets                                                \\ \hline
\citet{ravi-etal-2023-happens}                     & ECB+                                                    \\ \hline
\citet{ahmed-etal-2023-2}                                 & ECB+,GVC                                                \\ \hline
\citet{eirew-etal-2022-cross}                      & WEC                                                     \\ \hline
\citet{hsu-horwood-2022-contrastive}               & ECB+                                                    \\ \hline
\citet{bugert-gurevych-2021-event}                 & Hyperlink                                               \\ \hline
\citet{bugert-etal-2021-generalizing}              & \begin{tabular}[c]{@{}l@{}}ECB+,GVC,\\ FCC\end{tabular} \\ \hline
\citet{poumay-ittoo-2021-comprehensive-comparison} & ECB+                                                    \\ \hline
\citet{held-etal-2021-focus}                       & \begin{tabular}[c]{@{}l@{}}ECB+,GVC,\\ FCC\end{tabular} \\ \hline
\citet{caciularu-etal-2021-cdlm-cross}             & ECB+                                                    \\ \hline
\citet{cattan2021cross}                            & ECB+                                                    \\ \hline
\citet{eirew-etal-2021-wec}                        & WEC, ECB+                                               \\ \hline
\citet{allaway-etal-2021-sequential}               & ECB+                                                    \\ \hline
\citet{meged-etal-2020-paraphrasing}               & ECB+                                                    \\ \hline
\citet{zeng-etal-2020-event}                       & ECB+                                                    \\ \hline
\citet{barhom-etal-2019-revisiting}                & ECB+                                                    \\ \hline
\citet{choubey-huang-2017-event}                   & ECB+                                                    \\ \hline
\citet{upadhyay-etal-2016-revisiting}              & ECB+                                                    \\ \hline
\citet{yang-etal-2015-hierarchical}                & ECB+                                                    \\ \hline
\end{tabular}
\caption{\label{table:cdcr-paper-datasets}
Cross-document event coreference resolution papers and the corresponding datasets.  Most papers only evaluate one dataset to validate their claims.
}
\end{table}

Despite progress in ECR, two key challenges persist. Firstly, the evaluation of ECR study is often limited to a small number of datasets, hindering the assessment of the generalizability. To shed light on this issue, we deliver a thorough analysis of 39 papers from reputable NLP conferences (ACL, EMNLP, NAACL, EACL, and COLING) and journals (TACL and CL) published between 2015 and 2023.  Among these papers, 18 focused on CDCR, while 21 focused on WDCR.  In Table~\ref{table:cdcr-paper-datasets}, we present the findings related to CDCR papers. Notably, although there are 5 publicly available CDCR datasets, namely ECB+~\citep{cybulska-vossen-2014-using}, GVC~\citep{vossen-etal-2018-dont}, FCC~\citep{bugert2020breaking}, Hyperlink~\citep{bugert-gurevych-2021-event}, and WEC~\citep{eirew-etal-2022-cross}, 14 out of 18 CDCR-related papers only utilize a single dataset to validate their claims.  Similar observations can be made for WDCR papers and their respective datasets, which are detailed in Table~\ref{table:wdcr-paper-datasets} (Appendix~\ref{sec:appendix1}). While the need for multi-dataset evaluation has been acknowledged by~\citet{bugert-etal-2021-generalizing}, the complexity of reproducing state-of-the-art models across datasets remains a barrier. 

\begin{figure*}
	\centering
	\includegraphics[width=\textwidth]{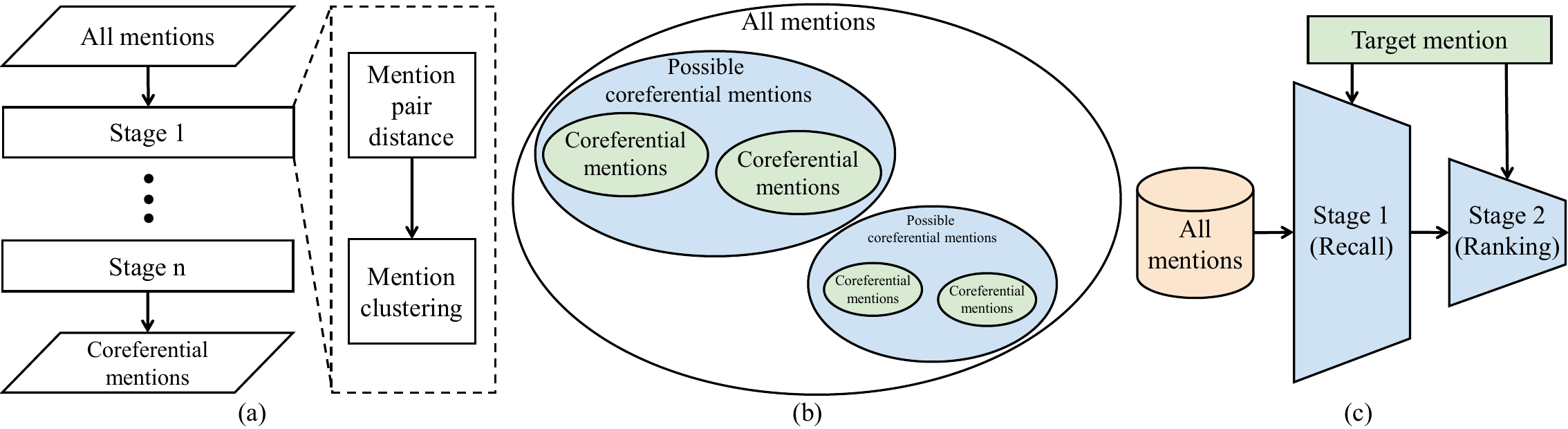}
	\caption{The ECR task is accomplished through a pipeline comprising multiple models. Here is a concise breakdown: (a) ECR pipeline: The pipeline consists of multiple stages, each comprising two steps: computing mention pair distances and clustering mentions. Each step involves a specific model.  (b) Illustration of a two-stage ECR pipeline: In the first stage, mentions are divided into clusters, where co-referential event mentions are likely. The second stage further divides these clusters into smaller ones, indicating coreferential event mentions.  (c) Finding coreferential mentions: This process resembles item recommendation in recommender systems~\citep{10.1145/2959100.2959190,ijcai2022p0771}. It involves identifying coreferential mentions for a given mention by progressing through the pipeline stages.
\label{fig:ecr-pipeline-model-vs-rec}
}
\end{figure*}

Secondly, ECR typically employs a multi-stage pipeline as depicted in Figure~\ref{fig:ecr-pipeline-model-vs-rec}. The reason for utilizing a pipeline approach in ECR is that it necessitates the computation of distances between all pairs of event mentions.  Employing a complex model to predict distances for such an enormous number of mention pairs (which is proportional to the square of the number of event mentions) is infeasible.  Therefore, similar to recommender systems~\citep{10.1145/2959100.2959190,ijcai2022p0771}, an ECR pipeline comprises multiple stages.  Initially, documents are grouped into clusters using straightforward strategies, such as document pre-clustering or topic priors. Only event mentions within the same cluster are considered potentially coreferential. Then, advanced models are employed to estimate the distances between event mention pairs within the same cluster, thereby substantially narrowing down the pool of pairs that require evaluation. In the progressively smaller and more specific clusters described above, event mentions are considered to be coreferential. However, variations in the number and type of models used across pipelines make it difficult to determine their comparative effectiveness. For instance, the CDCR pipeline LH+D$_{small}$ consists of three stages~\citep{ahmed-etal-2023-2} while the pipeline proposed by~\citet{held-etal-2021-focus} comprises only two stages, corresponding to the second and third stages in LH+D$_{small}$. Moreover, the models used for these phases are different. Additionally, due to its nature as a system rather than a single model, reproducing an ECR pipeline is more challenging.

To address these issues, we conduct a comprehensive investigation by reproducing 7 representative ECR pipelines, including 2 WDCR and 5 CDCR pipelines. Additionally, we rigorously evaluate these pipelines on 10 popular datasets, providing valuable baselines for future research and development. We ensure fair comparisons by evaluating the models used in different stages under consistent settings. To further facilitate the implementation and evaluation of new pipelines and assess their models on diverse datasets, we have developed and made available an open-source library called EasyECR. This library not only provides a unified data structure for ECR datasets but also offers a clear and intuitive abstraction of the ECR modeling process. Furthermore, the reproduced pipelines and utilized datasets are seamlessly integrated into the EasyECR, enabling researchers to readily explore and extend the field of ECR.

To summarize, our contributions are two-fold: (\uppercase\expandafter{\romannumeral1}) To make it easy to evaluate ECR pipelines on multiple datasets and compare models in ECR pipelines fairly, we develop EasyECR, the first unified library for implementing and evaluating ECR pipelines, incorporating seven representative pipelines and ten popular datasets. (\uppercase\expandafter{\romannumeral2}) Extensive experiments are conducted on multiple datasets using EasyECR. The experimental results provide valuable baselines for future research and models in different ECR pipelines are compared fairly.

\section{Related Work}
ECR is the task of clustering event mentions that refer to the same real-world event~\citep{lu2018event, liu2023brief}, which includes two subtasks: WDCR and CDCR. Numerous studies have been conducted on both WDCR and CDCR. In the following sections, we will provide an overview of these studies.

\textbf{WDCR} focuses on clustering event mentions within the same document~\citep{xu-etal-2022-improving,xu-etal-2023-corefprompt}. In recent studies, two noteworthy WDCR models have been reproduced. GLT~\citep{xu-etal-2022-improving} goes beyond previous methods by incorporating a Longformer-based encoder to capture document-level embeddings.  It also introduces an event topic generator to infer latent topic-level representations.  By combining sentence-level embeddings, document-level embeddings, and topic-level representations, GLT enhances coreference prediction.  CorefPrompt~\citep{xu-etal-2023-corefprompt} addresses the issue of coreference judgment reliance on event mention encoding.  It transforms ECR into a cloze-style masked language model (MLM) task, enabling simultaneous event modeling and coreference discrimination within a single template, leveraging a fully shared context.  Since the datasets used by GLT and CorefPrompt (KBP 2015, KBP 2016, and KBP 2017) contain short documents with a limited number of event mention pairs, both models consist of a single stage.  However, it is important to note that GLT and CorefPrompt are evaluated on only one dataset (KBP 2017), and their generalization performance on other datasets remains unknown.

\textbf{CDCR} involves clustering event mentions that may originate from the same document or different documents~\citep{lu2018event}.  Due to the vast number of event mention pairs in the CDCR task, it is typically addressed through multi-stage pipelines.  One example is the pipeline proposed by~\citet{held-etal-2021-focus}, which comprises two stages. The first stage utilizes a simple model to generate representations for event trigger and arguments, identifying potential coreferential mentions using these representations. The second stage utilizes a complex model to predict whether mentions and their top-k neighbors are coreferential. Another pipeline proposed by~\citet{hsu-horwood-2022-contrastive} also consists of two stages. In the first stage, mentions are separated by clustering documents, while the second stage employs contrastive representation learning to generate event mention representations, which are then used for clustering.  LH+D$_{small}$~\citep{ahmed-etal-2023-2} is a three-stage CDCR pipeline.  However, the different CDCR pipelines employ multiple models, making it unclear which model perform best in similar roles. 

\begin{figure*}
	\centering
	\includegraphics[width=\textwidth]{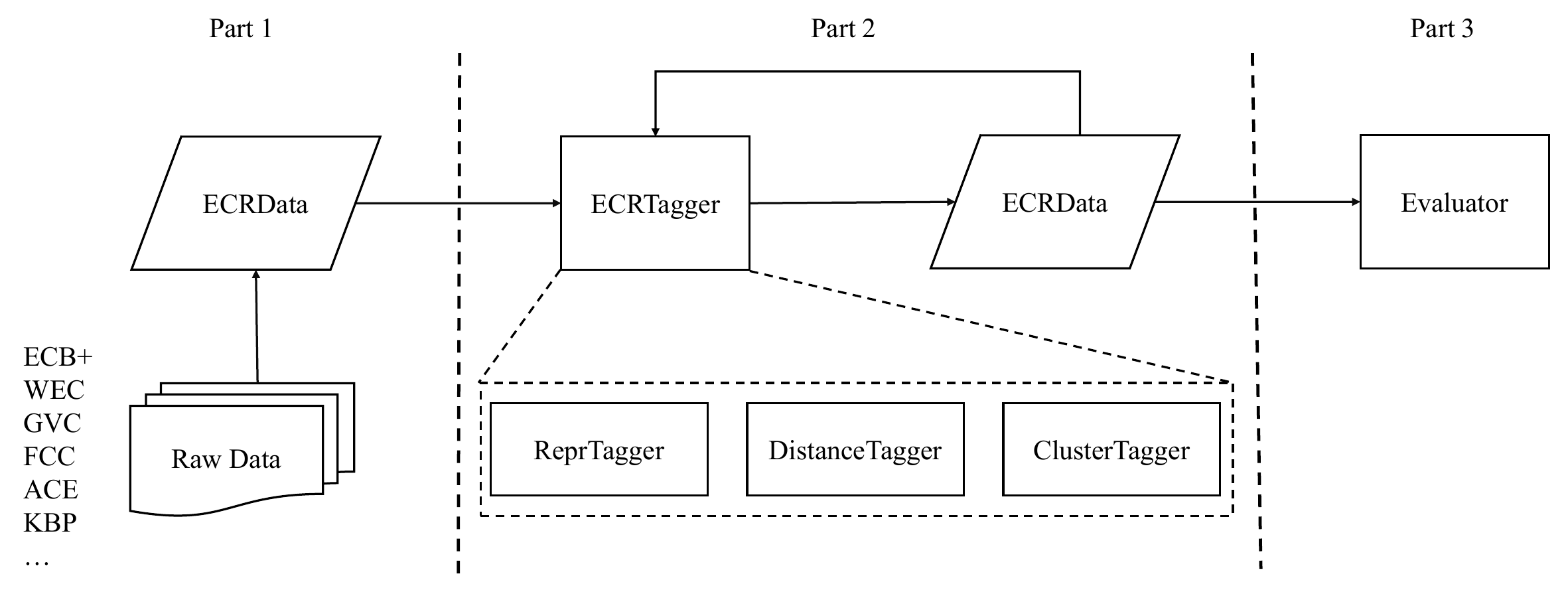}
	\caption{The framework of EasyECR. The Part 2 includes a loop corresponding to Figure~\ref{fig:ecr-pipeline-model-vs-rec} (a).
 \label{fig:ecr-pipeline}}	
\end{figure*}


\section{EasyECR}

EasyECR, as depicted in Figure~\ref{fig:ecr-pipeline}, consists of three key components: ECRData, ECRTagger, and Evaluator.  ECRData serves as a standardized data structure for ECR datasets within the EasyECR library.  ECRTagger assigns unique tags to individual event mentions in ECRData, with event mentions sharing the same cluster tag considered coreferential. Evaluator measures the performance of an ECR pipeline by comparing the event mention tags.  In the subsequent sections, we will provide a comprehensive description of each of these components.

\subsection{ECRData}
EasyECR incorporates three fundamental entities: document, mention, and ECRData, which collectively facilitate event coreference resolution.  A document corresponds to a textual passage, while a mention represents an event-related expression within a document.  Typically, a mention comprises an event trigger, which is the most explicit word or phrase indicating the occurrence of an event. Additionally, a mention may contain supplementary event details, such as event type and arguments. For example, as illustrated in Figure~\ref{fig:example}, ``saying'', ``negotiations'', and ``direct'' are event triggers in the mentions of ``THR is saying'', ``director Gary Ross is still in negotiations'', and ``to direct the sequel'', respectively.

ECRData encompasses a collection of documents and event mentions, with each entity possessing a unique identifier. Each mention is associated with a document via a document ID and linked to a corresponding event through an event ID. Mentions with the same event ID are considered to be coreferential. The flexible ECRData structure accommodates various existing ECR datasets, facilitating comprehensive coverage and compatibility.  


\subsection{ECRTagger}
In our EasyECR, an ECRTagger is a crucial component of the ECR pipeline that assigns tags to event mentions. Up to now, we have implemented three types of taggers.
\begin{compactitem}[-]
\item \textbf{ReprTagger}: This tagger generates representations for event mentions based on the event trigger, context, and other event information. A model proposed by~\citet{hsu-horwood-2022-contrastive} falls under this category.
\item \textbf{DistanceTagger}: This tagger predicts the distances (the distance here can commonly be defined as the difference between 1 and the probability that the two mentions refer to the same event) between event mention pairs using the event trigger, context, other event information, and mention representations generated by the ReprTagger.  Based on these distances, each event mention is assigned a tag containing its distances from other mentions.  Both the lemma distance model and the discriminator used by~\citet{ahmed-etal-2023-2} can be classified as DistanceTaggers.
\item \textbf{ClusterTagger}: This tagger clusters event mentions either based on their representations or the distances between mention pairs. Each event mention is assigned a cluster ID, and mentions with the same cluster ID are considered coreferential. Two ClusterTaggers have been implemented: ConnectedComponent, as used by~\citet{ahmed-etal-2023-2}, and the agglomerative clustering algorithm employed by \citet{ravi-etal-2023-happens}.
\end{compactitem}
There is no limitation on the number of taggers that can be applied to ECRData. A stage in an ECR pipeline can consist of either two taggers (ReprTagger and ClusterTagger) or three taggers (ReprTagger, DistanceTagger, and ClusterTagger). Importantly, the subsequent DistanceTagger only needs to predict distances for event mention pairs formed by mentions within the same clusters generated by the preceding ClusterTagger. This flexibility enables EasyECR to easily implement multi-stage ECR pipelines.

\subsection{Evaluator}
EasyECR incorporates the widely used coreference evaluation package, coval~\citep{moosavi-strube-2016-coreference}, for evaluating ECR performance.  The code has been slightly modified to support direct evaluation of ECR pipeline performance using the ECRData. This package has provided support for common evaluation metrics, including MUC~\citep{vilain1995model}, B-cubed~\citep{bagga1998algorithms}, CEAFe~\citep{luo2005coreference}, LEA~\citep{moosavi-strube-2016-coreference}, and CoNLL F1.  CoNLL F1, which is the average of MUC F1, B-cubed F1, and CEAFe F1, offers a comprehensive assessment of ECR. Therefore, we report only the CoNLL F1 score in the experiments, as done by~\citet{ahmed-etal-2023-2}.

\section{Experiments}

\subsection{Datasets}
We collect 10 commonly used datasets for our evaluation, which include: (1) Five datasets specifically designed for WDCR: ACE 2005 (English only)~\citep{Walker2006ace2005}, KBP~2015~\citep{mitamura2015overview}, KBP~2016~\citep{DBLP:conf/tac/MitamuraLH16}, KBP~2017~\citep{Mitamura2017EventsDC} and MAVEN-ERE~\citep{wang-etal-2022-maven}; (2) Five datasets for CDCR: ECB+~\citep{cybulska-vossen-2014-using}, GVC~\citep{vossen-etal-2018-dont}, FCC~\citep{bugert2020breaking}, FCCT~\citep{bugert-etal-2021-generalizing} and WEC~\citep{eirew-etal-2022-cross}.  The statistics of these datasets are reported in Table~\ref{table:dataset-statistics} and split statistics are shown in Table~\ref{table:dataset-statistics-more} (Appendix~\ref{sec:appendix2}).


\begin{table}
    \begin{center}
        \resizebox{\linewidth}{!}{
            \begin{tabular}{l|lll|l}
            \hline
            Dataset   & \#D   & \#M    & \#E   & C/W \\ \hline
            ACE 2005  & 535   & 4277   & 3341  & W   \\
            KBP       & 984   & 25058  & 16450 & W   \\
            MAVEN-ERE & 4480  & 133205 & 84285 & W   \\
            ECB+      & 976   & 6833   & 2741  & C   \\
            FCC       & 451   & 2618   & 217   & C   \\
            FCCT      & 451   & 3223   & 514   & C   \\
            GVC       & 510   & 7298   & 1413  & C   \\
            WEC       & 37129 & 43672  & 7597  & C   \\ \hline
            \end{tabular}
        }
    \end{center}
    \caption{\label{table:dataset-statistics}
Statistics of datasets incorporated into EasyECR. \#D, \#M, and \#E represents the number of documents, mentions, and events, respectively. C and W stand for CDCR and WDCR, respectively. KBP includes KBP 2015, 2016 and 2017.
}
\end{table}

All of these datasets have been seamlessly integrated into EasyECR.  By providing a directory as input, EasyECR generates an ECRData object that contains the dataset.  The directory structure and file formats remain the same as when the dataset is downloaded from a public source, requiring no modifications. This approach offers the advantage that even for non-free datasets like ACE 2005 that cannot be directly released, users can still effortlessly load and utilize them through EasyECR.  

\subsection{Reproduced Pipelines}
We have replicated seven notable representative pipelines in our comparison, encompassing two WDCR pipelines: GLT~\citep{xu-etal-2022-improving} and CorefPrompt~\citep{xu2023corefprompt}, as well as five  CDCR pipelines: DCT~\citet{held-etal-2021-focus}
CDCR-E2E~\citep{cattan2021cross}
CRL~\citep{hsu-horwood-2022-contrastive}, Lemma~\citep{bugert-etal-2021-generalizing,eirew-etal-2022-cross,ahmed-etal-2023-2} and
LemmaECR~\citep{ahmed-etal-2023-2}.  Lemma serves as a commonly used baseline and is also employed in the initial stage of LemmaECR~\citep{ahmed-etal-2023-2}. For all pipelines, we follow the settings in original papers.

Different models utilize distinct types of event information for event coreference prediction.  The inclusion of a dataset in model experiments depends on whether the dataset satisfies the specific event information requirements of the model.


\begin{table}
\centering
    \begin{tabular}{l|ll}
    \hline
    Method                       & Source   & CoNLL F1          \\ \hline
    \multirow{2}{*}{GLT}         & Original & \textbf{54.20} \\
                                 & EasyECR  & 52.31          \\ \hline
    \multirow{2}{*}{CorefPrompt} & Original & 54.23          \\
                                 & EasyECR  & \textbf{56.01} \\ \hline
    \end{tabular}
    \caption{\label{table:Reproduced Results-WDCR}
    Comparison between the results of reproduced WDCR pipelines and the results reported in original papers. All experiments are conducted on KBP 2017.
}
\end{table}

\subsection{Reproduced Results}
Table~\ref{table:Reproduced Results-WDCR} presents a comparison of the performance of the reproduced WDCR pipelines with the results reported in the original papers. Both GLT and CorefPrompt models are trained on the KBP 2015 and KBP 2016 datasets, and their evaluation is conducted on the KBP 2017 dataset.  It is observed that the performance of GLT is slightly lower than the results reported in the original paper, while the opposite is true for CorefPrompt.  The differences in performance for both models are approximately 2\%.  We are confident in the replication of these models, and a 2\% performance difference is considered reasonable.


\begin{table}
\centering
    \begin{tabular}{ccc}
    \hline
    \multicolumn{1}{c|}{Method}                    & Source   & CoNLL F1         \\ \hline
    \multicolumn{1}{c|}{\multirow{2}{*}{DCT}}      & Original & \textbf{85.70} \\
    \multicolumn{1}{c|}{}                          & EasyECR  & 73.07          \\ \hline
    \multicolumn{1}{c|}{\multirow{2}{*}{CDCR-E2E}} & Original & \textbf{81.00} \\
    \multicolumn{1}{c|}{}                          & EasyECR  & 78.40          \\ \hline
    \multicolumn{1}{c|}{\multirow{2}{*}{CRL}}      & Original & \textbf{81.80} \\
    \multicolumn{1}{c|}{}                          & EasyECR  & 72.08          \\ \hline
    \multicolumn{1}{c|}{\multirow{2}{*}{Lemma}}    & Original & 76.40          \\
    \multicolumn{1}{c|}{}                          & EasyECR  & \textbf{77.33} \\ \hline
    \multicolumn{1}{c|}{\multirow{2}{*}{LemmaECR}} & Original & \textbf{80.30} \\
    \multicolumn{1}{c|}{}                          & EasyECR  & 78.31          \\ \hline
    \end{tabular}
    \caption{\label{table:Reproduced Results-CDCR}
    Comparison between the results of reproduced CDCR pipelines and the results reported in original papers. All experiments are conducted on ECB+. 
}
\end{table}

Table~\ref{table:Reproduced Results-CDCR} compares the performance of reproduced CDCR pipelines with the results reported in the original papers. While Lemma, CDCR-E2E, and LemmaECR have been successfully reproduced, other models demonstrate lower performance.  Multi-stage pipelines, such as DCT and CRL, face additional complexities, making their reproduction more challenging.  For example, DCT utilizes a deep cluster model in its first stage, which generates representations and identifies potential coreferential mentions.  Reproducing the performance of this unsupervised and complex model is more difficult compared to simpler models like Lemma.  Additionally, unclear descriptions of CRL's sample generation strategy may have led to the omission of crucial details during reproduction.  Further analysis will be conducted to investigate the reasons behind the lower performance of the implemented DCT model.


\begin{table}
    \begin{center}
        \resizebox{\linewidth}{!}{
            \begin{tabular}{c|cccc}
            \hline
            \multirow{2}{*}{ID} & \multirow{2}{*}{Clustering} & \multirow{2}{*}{Subtopic} & \multirow{2}{*}{\begin{tabular}[c]{@{}c@{}}Keep\\ Singleton\end{tabular}} & \multirow{2}{*}{\begin{tabular}[c]{@{}c@{}}CoNLL\\ F1\end{tabular}} \\
                                &                             &                           &                                                                            &                                                                     \\ \hline
            1                   &                             & \Checkmark                & \Checkmark                                                                 & \textbf{77.3}                                                       \\ \hline
            2                   & \Checkmark                  &                           &                                                                            & 65.5                                                                \\ \hline
            3                   &                             & \Checkmark                &                                                                            & 66.3                                                                \\ \hline
            4                   &                             &                           & \Checkmark                                                                 & 71.7                                                                \\ \hline
            5                   &                             &                           &                                                                            & 59.6                                                                \\ \hline
            \end{tabular}
        }
    \end{center}
    \caption{\label{table:Impact-of-Evaluation-Settings}
Experimental results of Lemma on ECB+ with various evaluation settings:  The term ``Clustering'' refers to whether document clustering is utilized as a stage in the pipeline.  ``Subtopic'' indicates whether the subtopics provided by the dataset are used as the output of document clustering.  ``KeepSingleton'' denotes whether the singletons of event mentions are retained during pipeline evaluation.
}
\end{table}

\section{Analysis}
\noindent{\bf Impact of Evaluation Settings.}  To investigate the impact of different evaluation settings on the performance of CDCR pipelines, we conduct experiments specifically focusing on Lemma using the ECB+ dataset.  Table~\ref{table:Impact-of-Evaluation-Settings} displays the results obtained under various evaluation settings.  From the table, two conclusions can be drawn.  Firstly, both the Clustering and KeepSingleton settings significantly improve the CoNLL F1 score. Secondly, the Clustering and Subtopic settings demonstrate similar effects. These conclusions align with previous studies~\citep{cattan2021cross,bugert-etal-2021-generalizing}, which recommend the use of the ID 5 setting.  However, this recommendation has not been widely adopted.  In our comparison, we follow their suggestion and employ the ID 5 setting in Table~\ref{table:Impact-of-Evaluation-Settings} for evaluation, which may result in lower reported results in the subsequent analyses.

\begin{table}
    \begin{center}
        \resizebox{\linewidth}{!}{
            \begin{tabular}{c|c|c|c}
            \hline
            Method      & ACE 2005       & KBP2017        & MAVEN-ERE      \\ \hline
            GLT         & \textbf{64.64} & 52.31          & \textbf{64.34} \\
            CorefPrompt & 56.50          & \textbf{56.01} & 64.28          \\ \hline
            \end{tabular}
        }
    \end{center}
    \caption{\label{table:Generalization-Performance-of-ECR-Pipelines-WDCR}
Our experimental results of WDCR pipelines on multiple datasets in terms of CoNLL F1.
}
\end{table}

\begin{figure*}
	\centering
	\includegraphics[scale=0.3]{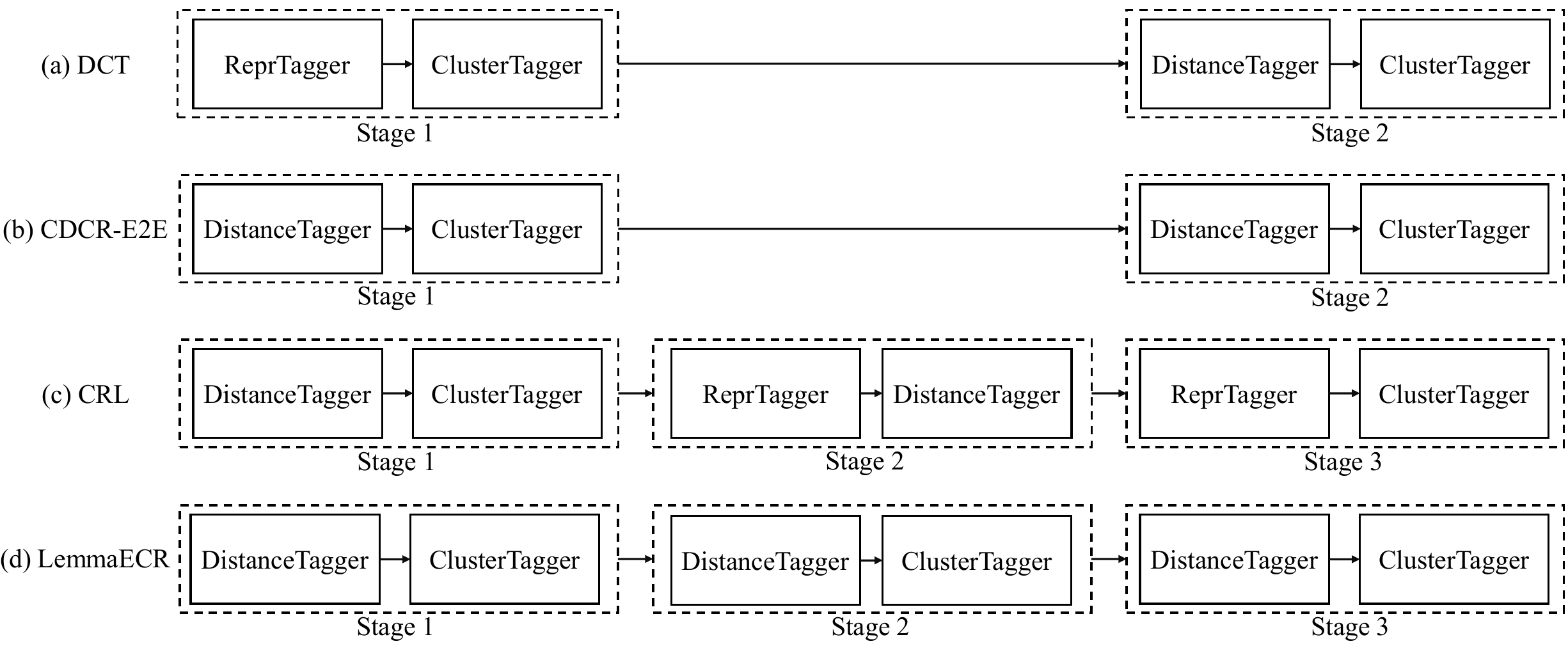}
	\caption{An illustration showing how four CDCR pipelines are constructed using ECRTaggers.}
	\label{fig:ecr-pipeline-instances}
\end{figure*}

\noindent{\bf Generalization.} We evaluate the implemented ECR pipelines on multiple datasets to showcase their generalization.  The experimental results of WDCR pipelines are presented in Table~\ref{table:Generalization-Performance-of-ECR-Pipelines-WDCR}, while the experimental results of CDCR pipelines are shown in Table~\ref{table:Generalization-Performance-of-ECR-Pipelines-CDCR}.  From the WDCR experimental results, we observe that CorefPrompt outperforms GLT on the KBP 2017 dataset, which aligns with the findings reported in~\citep{xu2023corefprompt}.  However, on the ACE 2005 dataset and the MAVEN-ERE dataset, the situation is reversed.  This indicates that the conclusions drawn from the KBP 2017 dataset may not directly apply to other datasets, highlighting the importance of evaluating WDCR pipelines across multiple datasets. Similar conclusions can be drawn from Table~\ref{table:Generalization-Performance-of-ECR-Pipelines-CDCR}. In addition, in Table~\ref{table:Generalization-Performance-of-ECR-Pipelines-CDCR}, except for CDCR-E2E and Lemma, results from other pipelines on the WEC dataset are not available, due to the enormous number of mentions in the WEC dataset. These pipelines could not complete their experiments on the WEC dataset within an acceptable timeframe.


\begin{table}
    \begin{center}
        \resizebox{\linewidth}{!}{
            \begin{tabular}{l|l|l|l|l}
            \hline
            Method   & ECB+          & WEC  & GVC           & FCCT          \\ \hline
            DCT      & 37.4          & -    & 29.7          & 34.3          \\
            CDCR-E2E & 59.6          & 56.0 & 32.5          & \textbf{44.4} \\
            CRL      & \textbf{67.7} & -    & 31.3          & 42.9          \\
            Lemma    & 59.6          & 40.7    & 30.5          & 37.8          \\
            LemmaECR & 65.9          & -    & \textbf{55.1} & 29.4          \\ \hline
            \end{tabular}
        }
    \end{center}
    \caption{\label{table:Generalization-Performance-of-ECR-Pipelines-CDCR}
Our experimental results of CDCR pipelines on multiple datasets in terms of CoNLL F1.
}
\end{table}

\noindent{\bf Comparison of Recall Models and Rank Models.}  Among the implemented pipelines, four CDCR pipelines consist of multiple stages, as illustrated in Figure~\ref{fig:ecr-pipeline-instances}.
\begin{compactitem}[-]
\item DCT~\citep{held-etal-2021-focus} consists of two stages. The first stage utilizes a deep cluster model (ReprTagger) to generate representations for event mentions and identify potential coreferential mentions.  The second stage trains a cross-encoder model (DistanceTagger) to directly predict the distances between event-mention pairs.    
\item CDCR-E2E~\citep{cattan2021cross} also includes two stages.  The first stage employs a DistanceTagger to compute distances between document pairs and a ClusterTagger to cluster the documents, with event mentions from the same document belonging to the same cluster.  The second stage trains a cross-encoder model (DistanceTagger) to predict distances between mention pairs.
\item CRL~\citep{hsu-horwood-2022-contrastive} comprises three stages. The first stage involves computing distances between document pairs using a DistanceTagger and clustering the documents using a ClusterTagger. The second stage employs a ReprTagger to generate event mention representations and a DistanceTagger to compute distances between event mention pairs based on these representations. The distances are then used to identify the most similar and least similar event mentions, generating positive and negative training samples for the third stage. The third stage trains a more effective ReprTagger.
\item LemmaECR~\citep{ahmed-etal-2023-2} also consists of three stages. The first stage computes distances between document pairs using a DistanceTagger and clusters the documents using a ClusterTagger.  The second stage uses a simple DistanceTagger to compute distances between event-mention pairs and clusters the event mentions accordingly, considering only the lemma of the event-mention trigger.  The second stage then trains a cross-encoder model (DistanceTagger) to predict distances between event-mention pairs.
\end{compactitem}
For convenience, similar to recommender systems~\citep{10.1145/2959100.2959190,ijcai2022p0771}, we categorize the models used in earlier stages of the ECR pipeline as recall models, while those used in later stages are referred to as rank models.  In DCT, the ReprTagger (Cluster) in the first stage, the ReprTagger (CRL) in the third stage of CRL, and the DistanceTagger (Lemma) in the second stage of LemmaECR are recall models. The DistanceTagger (DCT) in the second stage of DCT, the DistanceTagger (CDCR-E2E) in the second stage of CDCR-E2E, and the DistanceTagger (LemmaECR) in the third stage of LemmaECR are rank models.

\begin{table}[t]
\centering
    \begin{tabular}{c|c}
    \hline
    Method  & CoNLL F1          \\ \hline
    Cluster & 27.42          \\
    Lemma   & 59.60          \\
    CRL     & \textbf{67.68} \\ \hline
    \end{tabular}
    \caption{\label{table:Comparison-of-Earlier-Stage-ECRTaggers}
The performance of recall models on ECB+.
}
\end{table}

\begin{table}[t]
\centering
    \begin{tabular}{c|c|c}
    \hline
    Recall Model & Rank Model & CoNLL F1          \\ \hline
    Lemma           & DCT           & 60.09          \\
    Lemma           & CDCR-E2E      & \textbf{61.62} \\
    Lemma           & LemmaECR      & 60.01          \\ \hline
    \end{tabular}
    \caption{\label{table:Comparison-of-Later-Stage-ECRTaggers}
The performance of rank models on ECB+.
}
\end{table}

\begin{table}[t]
\centering
    \begin{tabular}{c|c}
    \hline
    ClusterTagger      & CoNLL F1          \\ \hline
    ConnectedComponent & \textbf{59.60} \\
    Agglomerative      & 27.42          \\ \hline
    \end{tabular}
    \caption{\label{table:Comparison-of-ClusterTaggers}
The experimental results of Lemma pipeline using different ClusterTaggers on the ECB+ dataset.
}
\end{table}

Table~\ref{table:Comparison-of-Earlier-Stage-ECRTaggers} presents the performance of three recall models, yielding three key observations: (1) There are notable performance differences among the models, highlighting the critical role of the recall model in overall pipeline performance. DCT's poor reproduction performance can be attributed to limitations in its recall model.  (2) While CRL demonstrates subpar reproduction performance under the commonly used evaluation setting, it surpasses state-of-the-art models in a more reasonable setting (67.68 vs. 62.67). (3) The recall model significantly impacts the overall effectiveness of the ECR pipeline by generating training samples for the rank model and defining the clustering scope in the subsequent stage. Our experiments reveal an interdependence between these factors. For example, in the Lemma experiment, optimizing hyperparameters improves results but reduces the number of training samples for the rank model (from nearly 30,000 to less than 10,000).  Hence, further exploration is necessary to comprehend the influence of the recall model on the overall performance of the ECR pipeline.


Table~\ref{table:Comparison-of-Later-Stage-ECRTaggers} presents the performance of three rank models for the Lemma  recall model.  The results show that the difference between the three rank models is small.

\noindent{\bf Comparison of ClusterTaggers.}  Different pipelines employ different ClusterTaggers. For instance, \citet{held-etal-2021-focus,ahmed-etal-2023-2} utilize the ConnectedComponent algorithm and \citet{cattan2021cross,hsu-horwood-2022-contrastive} employ the agglomerative clustering algorithm.  In this section, we conduct experiments to assess the impact of ClusterTaggers.  The experimental results of the Lemma pipeline using different ClusterTaggers are presented in Table~\ref{table:Comparison-of-ClusterTaggers}.  Notably, significant performance discrepancies exist among the different ClusterTaggers.  Therefore, for fair comparisons of other components within the ECR pipeline, it is essential to ensure that the ClusterTagger remains consistent.


\section{Conclusion}
In this paper, we develop EasyECR, an open-source ECR library designed to establish standardized data structures and enable fair comparisons among ECR pipelines. Extensive experiments are conducted using EasyECR and the results provide valuable baselines for future research. From the results, we draw two conclusions. First, the representative state-of-the-art ECR pipelines cannot generalize across multiple datasets, hence evaluating ECR pipelines on multiple datasets is necessary. Second, all models in ECR pipelines have a great effect on pipeline performance, therefore, when models corresponding to one step in ECR pipelines are compared, the other models need to remain identical.




\section{Limitations}
In ECR pipelines, preceding stages usually generate representations for event mentions and then find potential coreferential event mentions for target mentions based on the representations. Compared to predicting distances of event mention pairs using cross-encoder models~\citep{held-etal-2021-focus,ahmed-etal-2023-2}, computing the cosine or other distances between two mention representations is much faster. However, when the number of event mentions is larger, such as the WEC~\citep{eirew-etal-2022-cross} dataset, this step is still very time-consuming in EasyECR due to the simple vector search implementation. To mitigate this issue, FAISS~\citep{douze2024faiss,johnson2019billion} or other efficient vector search tools can be incorporated into EasyECR in the future. Additionally, our EasyECR has room for enhancement, with upcoming developments including an error-analysis module. This module will analyze instances where the model has incorrectly predicted event mentions as coreferential.

\bibliography{anthology,custom}
\bibliographystyle{acl_natbib}

\appendix

\section{WDCR Papers and The Corresponding Datasets}
\label{sec:appendix1}
Tabel~\ref{table:wdcr-paper-datasets} lists 21 WDCR papers and their corresponding datasets.  It is shown that most papers only evaluate a dataset to validate their claims. 

\begin{table*}
\centering
\begin{tabular}{|l|l|}
\hline
Paper                                 & Datasets                         \\ \hline
\citep{yao-etal-2023-learning}         & ACE2005                          \\ \hline
\citep{xu-etal-2023-corefprompt}       & KBP 2017 (trained on KBP 2015 and KBP 2016)        \\ \hline
\citep{wang-etal-2022-maven}           & MAVEN-ERE                        \\ \hline
\citep{xu-etal-2022-improving}         & KBP 2017 (trained on KBP 2015 and KBP 2016)         \\ \hline
\citep{lu-ng-2021-constrained}         & KBP 2017                          \\ \hline
\citep{kriman-ji-2021-joint}           & ACE05-E+                         \\ \hline
\citep{lai-etal-2021-context}          & ACE2005, KBP2016                 \\ \hline
\citep{choubey-huang-2021-automatic}   & KBP2017, RED                     \\ \hline
\citep{minh-tran-etal-2021-exploiting} & KBP2016, KBP2017 (trained on KBP 2015)         \\ \hline
\citep{lu-ng-2021-conundrums}          & ACE2005, KBP2017                 \\ \hline
\citep{joshi-etal-2019-bert}           & OntoNotes,GAP                    \\ \hline
\citep{huang-etal-2019-improving}      & KBP2017 English                  \\ \hline
\citep{lee-etal-2018-higher}           & OntoNotes 5.0                    \\ \hline
\citep{choubey-huang-2018-improving}   & KBP2016,2017                     \\ \hline
\citep{liu-etal-2018-graph}            & KBP2015                          \\ \hline
\citep{choubey-etal-2018-identifying}  & KBP2016                          \\ \hline
\citep{lu-ng-2017-joint}               & KBP2016 English and Chinese      \\ \hline
\citep{lu-etal-2016-joint}             & KBP2015 English, ACE2005 Chinese \\ \hline
\citep{peng-etal-2016-event}           & ACE2005, KBP2015                 \\ \hline
\citep{araki-mitamura-2015-joint}      & ProcessBank                      \\ \hline
\citep{chen-ng-2015-chinese}           & ACE2005                          \\ \hline
\end{tabular}
\caption{\label{table:wdcr-paper-datasets}
Within-document event coreference resolution papers and the corresponding datasets.  Most papers only evaluate a dataset to validate their claims.
}
\end{table*}

\section{More Dataset Statistics}
\label{sec:appendix2}
More dataset statistics are shown in Tabel~\ref{table:dataset-statistics-more}.

\begin{table*}
\centering
    \begin{tabular}{l|l|lll|l}
    \hline
    Dataset                    & Split    & \#D   & \#M   & \#E   & C/W \\ \hline
    \multirow{3}{*}{ACE 2005}  & Training & 428   & 2887  & 2237  & W   \\
                               & dev      & 53    & 647   & 514   & W   \\
                               & test     & 54    & 743   & 590   & W   \\ \hline
    \multirow{3}{*}{KBP}       & Training & 735   & 18880 & 12341 & W   \\
                               & dev      & 82    & 2182  & 1402  & W   \\
                               & test     & 167   & 3996  & 2707  & W   \\ \hline
    \multirow{3}{*}{MAVEN-ERE} & Training & 2913  & 73939 & 67984 & W   \\
                               & dev      & 710   & 17780 & 16301 & W   \\
                               & test     & 857   & 41486 & -     & W   \\ \hline
    \multirow{3}{*}{ECB+}      & Training & 574   & 3808  & 1527  & C   \\
                               & dev      & 196   & 1245  & 409   & C   \\
                               & test     & 206   & 1780  & 805   & C   \\ \hline
    \multirow{3}{*}{FCC}       & Training & 207   & 1195  & 115   & C   \\
                               & dev      & 117   & 535   & 47    & C   \\
                               & test     & 127   & 888   & 55    & C   \\ \hline
    \multirow{3}{*}{FCCT}      & Training & 207   & 1469  & 236   & C   \\
                               & dev      & 117   & 680   & 111   & C   \\
                               & test     & 127   & 1074  & 167   & C   \\ \hline
    \multirow{3}{*}{GVC}       & Training & 358   & 5313  & 991   & C   \\
                               & dev      & 78    & 977   & 228   & C   \\
                               & test     & 74    & 1008  & 194   & C   \\ \hline
    \multirow{3}{*}{WEC}       & Training & 34132 & 40529 & 7042  & C   \\
                               & dev      & 1194  & 1250  & 233   & C   \\
                               & test     & 1803  & 1893  & 322   & C   \\ \hline
    \end{tabular}
    \caption{\label{table:dataset-statistics-more}
Statistics of datasets incorporated into EasyECR. \#D, \#M, and \#E represent the number of documents, mentions, and events, respectively. C and W stand for CDCR and WDCR, respectively. KBP includes KBP 2015, 2016, and 2017, where KBP 2015 and 2016 are used to train models, while KBP 2017 is used to evaluate models.
}
\end{table*}

\end{document}